# Automatic Bug Detection in Games using LSTM Networks

Elham Azizi
*Ontario Tech University*
Oshawa, Ontario, Canada
elham.azizi@ontariotechu.ca

Loutfouz Zaman
*Ontario Tech University*
Oshawa, Ontario, Canada
loutfouz.zaman@ontariotechu.ca

*Abstract*—We introduced a new framework to detect perceptual bugs using a Long Short-Term Memory (LSTM) network, which detects bugs in video games as anomalies. The detected buggy frames are then clustered to determine the category of the occurred bug. The framework was evaluated on two First Person Shooter (FPS) games. Results show the effectiveness of the framework.

## I. INTRODUCTION

Game User Research (GUR) is one of the fields that is experiencing a recent transition due to the growth of AI. GUR is concerned with understanding and improving player experience by relying on the interdisciplinary knowledge of human-computer interaction (HCI), game design, and experimental psychology. A core component of GUR methodology is playtesting. Modern advances in GUR tools, such as gameplay analytics have helped to improve the quality of data collected through playtesting. Bugs can have a significant impact on the level of immersion and overall experience a player has in a video game. When players encounter bugs, it can disrupt the narrative flow, break immersion, and even hinder progress within the game [1]. Bugs can manifest in various ways, including graphical glitches, physics issues, AI errors, and gameplay imbalances. These problems can range from minor annoyances to game-breaking issues that prevent players from advancing or completing certain objectives. Before games are launched, the industry directs resources and puts effort to minimize the impact of bugs. However, significant bugs slip through even in AAA games at launch. The most notable example in the recent history is *Cyberpunk 2077*. One of the reasons for this is the sheer size and complexity of AAA games. As a result, playtesting these large and complex games using human testers is becoming unreasonable [2]. Moreover, traditional methods of bug detection, such as manual testing have been effective, but they can also be time-consuming and costly [3]. Furthermore, the sheer variety of bugs that can arise is vast, making it impractical to test a game comprehensively at every level, particularly for complex and expansive video games. The emergence of bugs can vary based on the level at which a bug occurs, these are: low-level, engine, application, perceptual, and behavioral [4]. Eight classes of bugs can occur: localization, save/load, network, UI, gameplay, audio, graphical, logical. Bugs can be identified with rule-based frameworks in low, engine, and application levels [5] [6] [7]. However, perceptual, and behavioral bugs cannot be detected by defining logical rules to infer issues trivially. Perceptual bugs in games refer to issues that affect the appearance or display of game elements while behavioral bugs affect the game logic or mechanics. These bugs can significantly impact the player's experience and can often be detected by playing the game and observing their behavior. The focus of our work is on perceptual and behavioral bugs, where we are concerned with detecting if the border between a normal bug-free, and abnormal buggy behavior in games is distinguishable.

With the rise of Artificial Intelligence (AI), game developers now have a new tool to detect bugs in their games [5] [8] [9]. In this work, we use Long Short-Term Memory (LSTM) [10], a deep neural network architecture, for our behavioral and perceptual bug detection. We frame the bug detection problem as Anomaly Detection, which allows integrating the framework to different video games easily. In the context of video analysis, anomaly detection involves identifying unusual or abnormal events, behaviors, or patterns that deviate from the expected norm [11]. We tested our approach on two 3D First Person Shooter (FPS) games. After finding an anomaly in a video frame, we specify the type of bug presented in the frame using the Density-Based Spatial Clustering of Applications with Noise (DBSCAN) algorithm. The key contributions of our work are a new general framework for perceptual and behavioral bug detection, and an introduction of a bug identification method using semi-supervised algorithms.

## II. RELATED WORK

Previous research has shown that AI has emerged as a powerful tool for detection and addressing of bugs in video games. With the increasing complexity of modern video games and the need for faster release cycles, traditional methods of bug detection, which were manual, have become more time-consuming and costly. AI-based approaches offer a promising alternative by automating the bug detection process and providing faster and more accurate results.

Automated game exploration and bug identification are the two main components of Automated Bug Detection (ABD). Automated game exploration has been gaining significant attention due to the recent advancements in Reinforcement Learning (RL). Y. Zheng et al. [6] introduced a new framework, *Wuji*, for game exploration and automated bug detection with the use of deep reinforcement learning (DRL). *Wuji* combines DRL and evolutionary multi-object optimization to find four categories of bugs: crash, stuck, logical, and gaming balance. RL was also used by Yavuz et al. [12] to test mobile apps automatically. This method used Q-learning to explore graphical user interfaces (GUIs) of a collection of applications to create their behavior models, which were then utilized to train for two optimization objectives. These were the number of crashes and activity coverage. However, bug detection methods in [6] and [12] use the rule-based approach which cannot detect high-level perceptual and behavioral bugs that players encounter. The focus of our work is to design new models to address these high-level bugs.

Bergdahl et al. introduced a self-learning game testing framework using DRL [5]. This system can navigate and take advantage of the mechanics of a game though RL, which is guided by a predefined reward signal. RL was also used by Gordillo et al. [13] to train agents capable of exploring a game automatically to further test it for bug detection purposes. Gordillo et al. evaluated the proposed approach on a 500 m × 500 m × 50 m 3D map that was designed for creating an elaborate navigation landscape. The focus of this work was to maximize the testing environment coverage with an AI agent. While the focus in ABD has been more on game exploration, Pfau et al. [14] also use RL in its architecture to introduce a new autonomous framework, where games can be played and bugs can be reported. In their work, Pfau et al. paid special attention to adventure games and used discrete RL where short-term and long-term memory were used in pair as a solving mechanic. The works above mostly work on game environment exploration, rather than bug detection itself. Our work, however, detects bugs based on human played datasets and works around maximizing the detection of buggy frames.

Focusing more on behavioral and perceptual bugs, *World of Bugs* (*WOB*) employs learning-based methods by leveraging the rendered scene viewed by players [15]. An open platform for ABD testing has been developed for 3D game environments. An AI agent explores and captures video frames from the game environment as the player would have seen the game world. To detect bugs in the captured frames, an auto-encoder model was used to detect bugs by projecting input into a lower dimension. In total, the authors identified ten distinct types of perceptual bugs, with the detection accuracy varying across these categories. This performance discrepancy stemmed from specific instances where the object's rear geometry was not properly rendered, causing the model to interpret the observations as ordinary. As a part of our work, we use the *WOB* dataset, but we test the model on a different dataset with more complex objects. Our framework focuses more on bug detection than game exploration in the ABD context. Also, our approach strives to solve the issue of bug detection where the presence of a buggy frame depends on the previous frames. In other words, the model is capable to detect buggy frames that cannot be detected if the previous frames are not considered. For example, in geometry clipping bugs, where the agent or player can walk through the objects, some gameplay frames may not reveal the bug. The observations can look normal as the object disappears from the view.

### III. DATA

In this work, we have analyzed two different datasets: one was from *World of Bugs* (*WOB*) [15], the other one we developed using an FPS game from *Echo+* [16]. Both games are of the FPS genre. The gameplay videos were recorded using the mp4 format.

*1) World of Bugs:* *WOB* is an open platform for testing automatic bug detection in 3D games. The datasets created in this platform are divided into 10 perceptual bug categories: texture missing, texture corruption, Z-fighting, Z-clipping, geometry corruption, screen tear, black screen, camera clipping, boundary hole, and geometry clipping. The dataset is captured and collected by an AI agent and then is divided into train and test categories. The training data contains 300k observations and actions from 60 episodes of bug-free play (normal, no bugs). The testing data contains 500k videos that are abnormal (have bugs), 50k examples for each class of bugs.

*2) Echo+:* The FPS game used in *Echo+* is a publicly available open-source Unity game. The game was modified to introduce perceptual bugs. Black screen, texture corruption, and boundary hole are the three types of bugs that were added to the game. To evaluate this game, we have recorded and collected gameplay from 25 players. The average length of the recordings is 2 minutes. Each player played multiple times and in total, we collected 300 minutes of gameplay. The buggy and bug-free videos were split into 70% for training 30% for testing.

### IV. IMPLEMENTATION

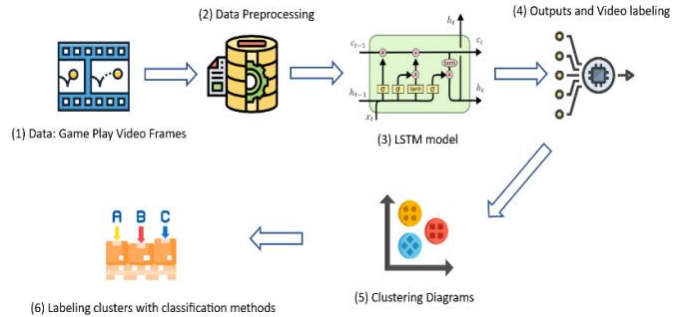

Figure 1. Framework architecture.

It is important for game developers to thoroughly test all possible scenarios and interactions to ensure that bugs are identified and fixed. Some bugs are independent from previous scenes and can occur randomly, while others are triggered by specific actions or events in previous frames. In these scenarios where one frame may look normal without considering the previous events in the game, the human tester or ABD system should consider the sequence of frames or actions yielding to the bug presence. To observe the sequence of actions, we have used the LSTM model in our framework. In LSTM, earlier inputs can affect later outputs in a sequence since these models can selectively forget or remember information over time. The LSTM architecture is a variant of recurrent neural networks (RNNs) designed to effectively model long-term dependencies in sequential data. It incorporates memory cells and gating mechanisms to regulate the flow of information. The memory cell stores and updates information over time, allowing the network to retain relevant information from previous time steps. The three key gates in an LSTM, namely the input gate, forget gate, and output gate, control the flow of information into and out of the memory cell. The input gate manages new information, the forget gate decides what to discard, and the output gate determines the output based on the current input and the cell state. Previously, creating tools like these centered around framework building or using detailed illustrations of the environment. This means they would be tightly integrated with the implementations of the games [6] [12]. Yet, in order to make the framework usable for all genres of games, as well as types of bugs, the problem is framed as Anomaly Detection. In video games, the emergence of bugs is an anomaly since these frames deviate significantly from the expected behavior of the system. Our framework identifies anomalies and compares each point in a frame to the normal data based on the time and the sequence in which they happened using an LSTM network. The overall architecture of the framework is illustrated in Figure 1.

The framework has been implemented on a Microsoft *Windows* 10 PC with Nvidia *GeForce* RTX 2080 Ti, with 16 GB RAM, and an AMD *Ryzen* 7 2700X CPU of 3.7 Ghz. During the training process, the amount of Video Random Access Memory (VRAM) available on the GPU determines how many frames or images can be loaded into memory at once. Because of the power limitations and memory shortages, we chose to work on grayscale frames. On the other hand, to solve the VRAM capacity problem, we chose to load 10 frames at a time into memory and process these frames as if they are a video clip having 10 frames in each time slot, so the window size is 10. The window slide technique determines the number of frames to be included in each section. The step size of the window slide is set to 1, which means we did not skip any frames. The step size can vary with the bug frequency emergence in a specific amount of time. For example, a step size of 2 means that between each two subsequent frames, one of them is skipped. This can be useful to save memory and time if a bug takes more than two frames to detect so that the model does not skip a bug. The step size of 1 is mandatory in games due to the sensitivity of bug occurrence in some scenarios, e.g., a black screen bug which can occur in 1 frame. Constructing a video clip of 10 frames makes it easier and faster to train and evaluate the model. At each iteration of constructing these clips, a data loader module calls for the address of frames that are saved in memory. The training on the two datasets followed the same model parameters and logic, however, the number of bug categories each of the datasets contains is different.

## V. EXPERIMENTAL RESULTS

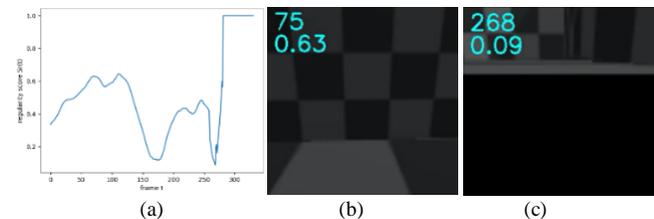

(a)           (b)           (c)

Figure 2. An example of the framework's output of a boundary hole bug in the *WOB* game. The captions in b) and c) represent the time and the RS of the frame. a) RS diagram, b) a normal frame, c) a buggy frame.

The framework was tested on *WOB* and *Echo+* FPS game datasets within their defined bug categories. To demonstrate the results more effectively, the model outputs a diagram for each video beside a caption on the video that shows the frame number and the Regularity Score (RS) of the corresponding frame after testing.

1) *WOB*: For each of the 10 bug categories, the RS diagram was created for every video frame. A sample example of the output with the related frame containing a boundary hole bug is demonstrated in Figure 2. As can be confirmed by the corresponding frame of the bug, which is 268, the RS value is dropping significantly to 0.09 because of the boundary hole bug.

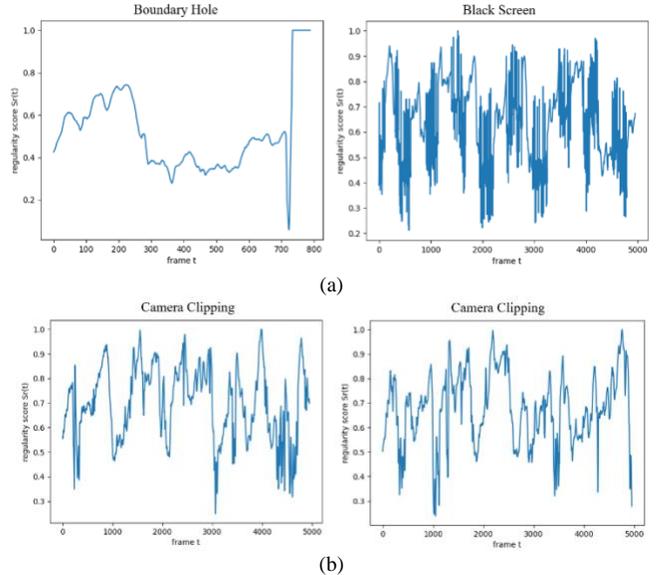

(a)

(b)

Figure 3. An illustration of RS diagram pattern of three bug categories.

We observed that in the *WOB* data the output pattern of the 10 categories of bugs was different from each other. However, the outputs of a category of a bug were similar to each other. Figure 3a demonstrates the results of the test on two different frames that have two different types of bugs: black screen and boundary hole. Figure 3b demonstrates the results of two frames that have the same type of bug: camera clipping. It can be inferred that the diagrams in Figure 3a follow different patterns of values and behavior. Meanwhile, the pattern in the diagrams of Figure 3b looks similar. Also, the observations from the diagrams show that the threshold value of RS for each type of bugs is distinct. The threshold of the RS refers to the value and the extent a bug occurred in the corresponding frame number.

To further extend our RS diagram similarity observations for bug categories, we used DBSCAN on the outputted diagrams, as it is well-suited for detecting dense and non-linear clusters in data. We then used pre-labeled diagrams, where their bug categories were defined, and compared the clusters with them. We then labeled each of the created clusters based on the pattern similarity they had with the pre-labeled diagrams. To evaluate the clustering performance, we used Homogeneity Score [17] which is a clustering evaluation that measures the degree to which each cluster contains only samples belonging to a single category of concepts. While other metrics like silhouette, and V-measure are indeed valuable for evaluating clustering algorithms, the specific context and goals of the study led to prioritize the homogeneity score as a relevant. Equation 1 illustrates the homogeneity score formula.

$$homogeneity_{score} = 1 - \frac{H(C|K)}{H(C)} \qquad (1)$$

where $H(C|K)$ is the conditional entropy of the class labels given the cluster assignments, and $H(C)$ is the entropy of the class labels.

The results showed that the algorithm was able to cluster diagrams with the homogeneity score of 0.74. The score, however, was improved to 0.85 when we removed two classes of diagrams of camera clipping and geometry clipping since these two classes outputted diagrams similar to each other.

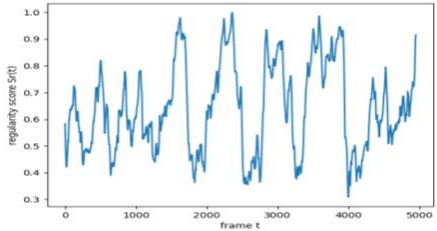

Figure 4. An RS diagram of texture corruption bug in the *Echo+* FPS game

2) After training the model with the bug free data in the *Echo+* FPS game dataset, we tested out the gameplays containing buggy frames. We observed that the model was able to detect anomalies in the gameplays but the diagrams were not following a similar pattern. This is because buggy videos of the *Echo+* dataset did not have similar bugs and the size of the testing data was not large. In contrast to *WOB*, we wanted to test if the model could detect anomalies in a more complex game environment containing objects and properties closer to an FPS game, including enemies, boss fight, and shooting. The test size was small since the game did not have too many bugs. However, the training data was large enough for the model to observe all possible actions or objects. Figure 4 illustrates a sample RS diagram output of a texture corruption bug category in one of the collected *Echo+* gameplays.

## VI. Limitation and Future Work

The framework has been tested on a game dataset played by human players, but there is uncertainty about its effectiveness in detecting bugs in highly complex games with larger worlds and diverse objects. While a well-trained model may be capable of handling such games, acquiring a suitable dataset for training will require the introduction of a new RL agent. However, limitations in training time and memory capacity are expected. In the future, we will test the algorithm on a more extensive game to assess its performance further. We will also test if testing the algorithm in a non-grayscale setting improves the results, especially in games where different objects are distinguished with colors.

## VII. Conclusion

Detection of perceptual and behavioral bugs is more challenging due to their complexity and the inability to define logical rules to distinguish them. In this work, we introduced a new framework to detect perceptual and behavioral bugs with the use of LSTM models. The bugs are defined as being anomalies in the gameplay, which are the incidents that the player does not expect to see or experience. The developer or tester can observe when and to what extent a bug has occurred using the RS metric. To specify the type of bugs occurred in frames with anomalies, we applied a clustering algorithm on RS diagrams with a semi-supervised learning method using DBSCAN. The clusters were labeled based on their similarity to pre-labeled RS diagrams. The framework was tested on two different 3D FPS games. The results showed that each two bugs from the same category followed a similar RS-time diagram pattern and each two buggy frames from different categories follow a distinct one. Our observations promise a new method for detecting perceptual and behavioral bugs using the introduced framework.

## Acknowledgment

We extend our gratitude to Daniel MacCormick for help with preparing the *Echo+* FPS game.